\documentclass[a4paper,twoside]{article}

\usepackage{epsfig}
\usepackage{subcaption}
\usepackage{calc}
\usepackage{amssymb}
\usepackage{amstext}
\usepackage{amsmath}
\usepackage{amsthm}
\usepackage{multicol}
\usepackage{pslatex}
\usepackage{apalike}
\usepackage{cite}
\usepackage{url} 
\usepackage{color,soul}
\usepackage{algorithm}
\usepackage{algorithmicx}
\usepackage{algpseudocode}
\usepackage{bm}
\usepackage{float}

\newcommand{\X}{\mathcal{X}}
\newcommand{\D}{\mathcal{D}}
\newcommand{\C}{\mathcal{C}}
\newcommand\Reals{\mathbb{R}}
\renewcommand{\vec}{\bm}
\newcommand{\layout}{\vec{\ell}}

\usepackage{SCITEPRESS}     

\begin{document}

\title{Optimizing Hospital Room Layout to Reduce the Risk of Patient Falls}

 \author{\authorname{Sarvenaz Chaeibakhsh\sup{1}, Roya Sabbagh Novin\sup{1}, Tucker Hermans\sup{2}, \\Andrew Merryweather\sup{1}, and Alan Kuntz\sup{2}}
\affiliation{\sup{1}Department of Mechanical Engineering, University of Utah, UT, USA}
 \affiliation{\sup{2}School of Computing, University of Utah, UT, USA}
 \email{sarvenaz.chaeibakhsh@utah.edu}
}


\abstract{Despite years of research into patient falls in hospital rooms, falls and related injuries remain a serious concern to patient safety.
In this work, we formulate a gradient-free constrained optimization problem to generate and reconfigure the hospital room interior layout to minimize the risk of falls.
We define a cost function built on a hospital room fall model that takes into account the supportive or hazardous effect of the patient’s surrounding objects, as well as simulated patient trajectories inside the room.
We define a constraint set that ensures the functionality of the generated room layouts in addition to conforming to architectural guidelines.
We solve this problem efficiently using a variant of simulated annealing. 
We present results for two real-world hospital room types and demonstrate a significant improvement of $18\%$ on average in patient fall risk when compared with a traditional hospital room layout and $41\%$ when compared with randomly generated layouts.
}

\onecolumn \maketitle \normalsize \setcounter{footnote}{0} \vfill

\section{\uppercase{Introduction}}
\label{sec:introduction}

\noindent Patient falls in healthcare settings have a severe impact on patient outcomes, resulting in increased morbidity, length of stay, and reduced quality of life.
Further, unnecessary falls incur significant financial costs to both patients and the healthcare system.
Every year in the US 700,000 to 1,000,000 people fall in hospitals \cite{hughes2008patient}.
Studies have shown that close to one-third of these falls are preventable \cite{cameron2012interventions}.

Substantial research studying hospital falls and related fall injuries has been done, yet the range of preventive and protective interventions is still somewhat limited.
Current preventive interventions mostly include  solutions to notify staff when the patient egresses, such as bed alarms and video monitoring, with the intention that the staff can intervene with the patient prior to a fall occurring \cite{alert2015preventing}\cite{callis2016falls}.
Safety measures have also been studied to reduce injury severity when a fall occurs, such as hip protectors and compliant flooring \cite{willgoss2010review}.
Yet fall rates continue to be unacceptably high and are even increasing \cite{hsiao2016fall}\cite{alert2015preventing}, representing a serious threat to patient safety.

\begin{figure}
    \centering
    \begin{subfigure}[t]{0.2\textwidth}
        \includegraphics[width=\textwidth]{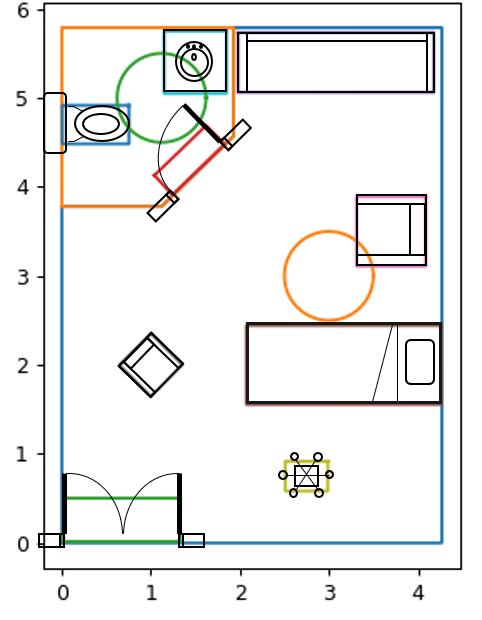}
        \caption{Traditional layout}
        \label{subfig: trad room}
    \end{subfigure}
    \begin{subfigure}[t]{0.2\textwidth}
        \includegraphics[width=\textwidth]{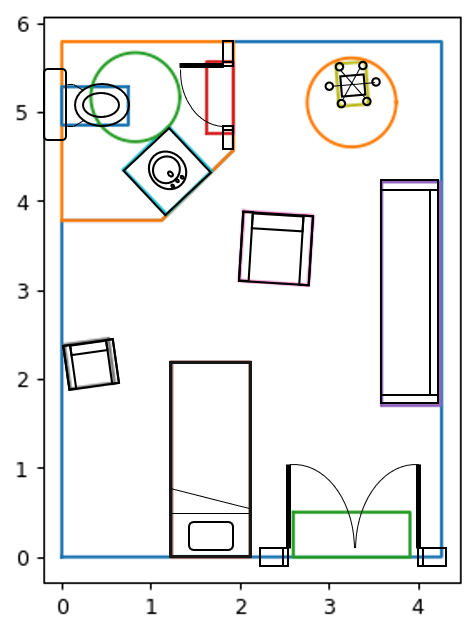}
        \caption{Optimized layout}
        \label{subfig: inbard room vs trad}
        \vspace{10pt}
    \end{subfigure}\qquad

    \begin{subfigure}[t]{0.2\textwidth}
        \includegraphics[width=\textwidth]{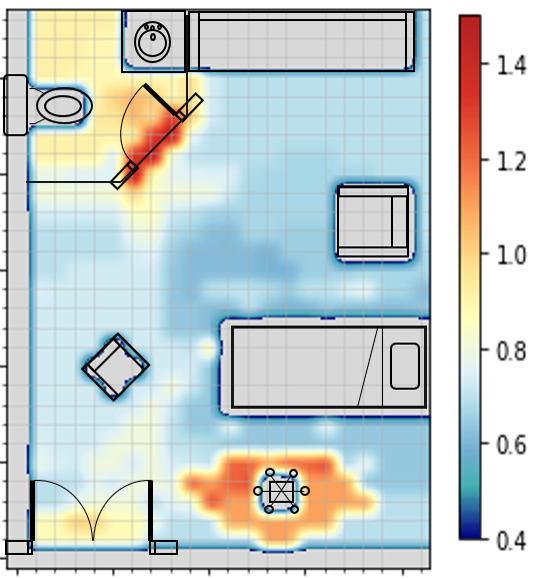}
        \caption{ Risk of fall heatmap for traditional layout}
        \label{subfig: trad HM}
    \end{subfigure}
    \begin{subfigure}[t]{0.2\textwidth}
        \includegraphics[width=\textwidth]{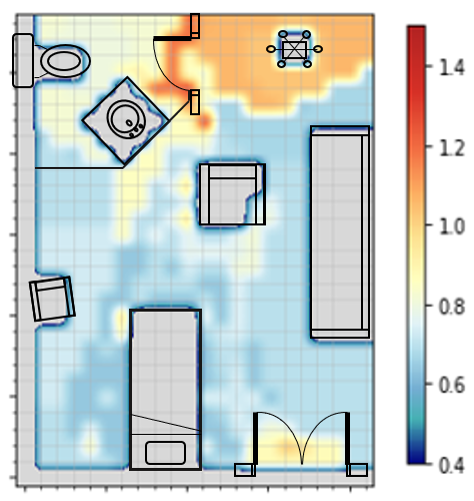}
        \caption{ Risk of fall heatmap for optimized layout}
        \label{subfig: inboard vs tradHM}
    \end{subfigure}
   \vspace{5pt}
    \caption{Traditional and optimized room layout evaluation with respect to fall risk. Figures (a) and (b) show the schematic of the rooms and figures (c) and (d) show the corresponding heat map of the risk of fall as evaluated by the fall risk model.}
    \label{fig:first}
\end{figure}

A variety of causes contribute to patient falls. However, environmental hazards and accidents have been identified as the most common causes of falls among the elderly \cite{joint2015preventing}.
Recently, a metric of patient fall risk during unassisted ambulation in a hospital room was proposed that considers the layout of the room~\cite{sabbagh2020development}.
This method estimates the fall risk of a room by considering room design factors including lighting, flooring type, door operation (swinging or sliding), and supporting objects (e.g., furniture, grab bars, bed rails, etc) in the room.

However, the relationship between the fall risk and the room layout as defined by the metric in \cite{sabbagh2020development} is complex and non-intuitive, making manually adjusting the layout to reduce the fall risk infeasible.

In this work, we build upon this fall model and employ gradient-free optimization to automatically generate hospital room layouts that reduce the risk of patient falls (see Fig.~\ref{fig:first}).
Specifically, we adapt simulated annealing to optimize features of multiple real-world hospital rooms including the placement of objects such as the patient bed, sofa, patient chair, IV pole, toilet and sink; the placement of lighting; and the locations of the main door and bathroom door.

We employ real-world architectural design guidelines as constraints, such as specific minimum clearance between sets of objects, which are representative of residential construction requirements to ensure room functionality \cite{ramsey2007architectural, neufert2012architects}.
Using these guidelines as constraints enables the method to take steps toward designing feasible rooms that reduce the risk of patient falls.

This paper represents the first work that optimizes hospital room layouts to reduce the risk of patient falls.

\section{\uppercase{Related Work}}
\noindent With patient falls being a serious issue in health care settings, many solutions have been proposed in the literature to reduce falls and post-fall injuries\cite{callis2016falls, clarke2012preoperative, mayo1994randomized}.
Our method is conceptually similar to computerized layout planning. The focus of our method is on leveraging optimization to reduce the risk of fall in hospital rooms by changing the layout of the furniture and medical devices to create safer surroundings and pathways for the patient. 
In this section, we discuss the history of patient falls and fall prevention methods and discuss existing applications of computerized layout planning in general and in health care. 

\subsection{Hospital Fall Prevention Strategies}\label{sec: background: fall prevention}
\noindent Despite extensive efforts to prevent patient falls, falls in acute care hospital rooms remain a serious issue\cite{hsiao2016fall}.
Around one-third of hospital falls result in injuries to the patient and over 84\% of adverse incidents in hospitals that lead to co-morbidity and mortality are associated with falls\cite{choi2011developing, aranda2013instruments}.
Most previous research on hospital falls has focused on the effect of intrinsic factors including medications, and less attention has been paid to extrinsic factors relating to the hospital room itself and the patient's physical environment \cite{callis2016falls}.
Different preventive strategies such as patient education \cite{clarke2012preoperative}, physical restraints, alarms \cite{mayo1994randomized, tideiksaar1993falls} and flooring \cite{donald2000preventing} have tried to address this serious issue.
Still, hospital falls continue to be the leading cause of injuries to the senior population in health care facilities \cite{joint2015preventing}.

The layout of furniture and medical equipment in hospital rooms has been shown to have a significant impact on the safety of patients \cite{hignett2006review}.
Studies such as \cite{joint2015preventing} and \cite{hignett2006review} highlight the significance of patients' physical surroundings in their safety and suggest several environmental strategies, such as ensuring adequate lighting and appropriate flooring types, to reduce the risk of fall.
In this work, we build on this concept; the physical layout including furniture and equipment in a hospital room can be optimized to improve patient safety and decrease the risk of falls.
We accomplish this using an optimization method in a manner conceptually similar to computerized layout planning.

\subsection{Computerized Layout Planning}\label{sec:background: CLP}
\noindent Computerized layout planning refers to leveraging computers in allocating space while a set of criteria and constraints are met and/or some objectives are optimized.
The demand for computerized layout planning has been growing since the 1960's when the first ideas for rule-based computerized layout planning began to take shape \cite{liggett1985optimal}.
Since then, much work has been done on computerized layout planning, including a few commercialized products.
Many of the commercialized computerized layout planning tools, such as Spacemaker~\cite{spaceMaker} and Planner 5D~\cite{Free3DHome}, focus on automatically planning the layout and space allocation for the placement of buildings rather than interior layout. These software packages use techniques in mathematical modeling, artificial intelligence, and architectural urban development to assist architects in designing multi-building residential sites and high level planning. 
Some of these methods generate different facility layouts based on predefined rules \cite{araghi2015exploring} while others use machine learning methods to design general building layouts \cite{merrell2011interactive}.
However, so far the application of machine learning methods have been limited in efficacy for large scale or detailed problems \cite{wu2018miqp, jamali2020review}.

Beyond space allocation and layout planning, where the concern is about the physical arrangement of objects and resources that consume space, computerized layout planning has also been applied to object placement in interior design.
In \cite{merrell2011interactive}, an interactive layout planner is proposed which takes an initial furniture arrangement and constraints and suggests new furniture configurations to the user. Gal et al. use a rule-based method to develop a framework to generate object layouts by solving a constraint satisfaction problem \cite{gal2014flare}. With advances in AI, virtual reality, and augmented reality, AI-based interior design tools are emerging.
The IKEA PLACE platform, developed by IKEA, virtually places the company's products in an area scanned via a cellphone camera \cite{IKEAApps55}.
Leaperr AI software \cite{leaperr} combines deep learning and image processing to suggest a design for the interior of a room based on a preference questionnaire filled out by the user.
Planner 5D is another AI-powered app that turns 2D blueprints into 3D and helps users visualize their desired furniture layout \cite{Free3DHome}.
The established AI-powered methods in general are designed to help a user design and visualize a space, but generally do so without any knowledge of design rules, layout constraints, or optimization techniques. 

Hospital department layout planning is one of the more focused research areas in computerized layout planning for health care facilities \cite{8627949, rismanchian2017process, lin2015integrating}.
In hospital department layout planning the location of hospital departments are rearranged to improve metrics such as patient travel time and relocation cost~\cite{jamali2020review}.
However, a method to inform the furniture layout in health care facilities to improve patient safety has yet to be studied.

In this paper, we address the placement of furniture, lighting, and doorways inside a hospital room to reduce patient falls.
The concept of satisfying constraints and optimizing the placement of objects exists in other application domains. Next we discuss two categories of existing methods that have been used to accomplish similar tasks in these domains: constructive (rule based) methods and iterative improvement (optimization-based) methods.

\subsubsection{Constructive/Rule Based Methods}\label{section: background Rule based model}
\noindent Constructive or rule-based methods build a constraint-satisfying solution by placing objects one-by-one in an iterative decision process.
Each object that is placed in the room has its own features that should be compatible with the previously placed items and the features of the specific environment, described as constraints.
For example, a bed and a sofa cannot be stacked, but a ceiling light fixture can be located above a bed.
Frequently, these research questions are framed as constraint satisfaction problems~\cite{ghedira2013constraint}, and employ backtracking methods~\cite{karumanchi2018algorithm} to solve problems with many complex constraints.

In \cite{tutenel2009rule}, the authors use a rule-based method to automatically create scenes for sand-box style video games. In \cite{merrell2011interactive}, an interactive layout planner is proposed which takes in an initial furniture arrangement and set of constraints, and suggests a new furniture configuration to the user.
These proposed layouts are based on architectural guidelines to ensure functionality as well as aesthetic appeal.
In \cite{gal2014flare} Gal uses a rule-based method to develop a framework to generate object layouts by solving a constraint satisfaction problem in placing objects in augmented reality.
However, these constructive models generally only produce a \emph{feasible} solution that satisfies a set of constraints, but do not consider producing an \emph{optimal} solution under some cost function.
In this paper, we adapt the concept of a constraint satisfaction problem to generate constraint-satisfying, feasible hospital rooms during our optimization process.

\subsubsection{Iterative Improvement Methods}\label{subsec: background-sa}
\noindent Iterative improvement, or optimization-based methods start with an initial solution and improve upon it over multiple iterations.
The initial solution is either generated randomly or can be introduced to the system by the user.
There are many types of iterative improvement optimization methods \cite{yang2014nature}.
Many iterative methods do not necessarily need gradient/derivative information, enabling them to solve a larger class of complex and discontinuous problems for which such gradient information may not be available.
Some iterative methods are inspired by natural processes, such as simulated annealing and genetic algorithms, and attempt to balance exploring unknown regions of a parameter space and exploiting existing knowledge in finding near-optimal solutions \cite{yang2014nature}.

Simulated annealing is a canonical probabilistic iterative improvement method that was first proposed by Kirkpatrick, Gelett, and Vecchi in 1983 \cite{kirkpatrick1983optimization}, further improved by {\v{C}}ern{\`y} in 1985~\cite{vcerny1985thermodynamical}, and has been the focus of much study since~\cite{bertsimas1993simulated,dowsland2012simulated, nikolaev2010simulated}.
This method, as indicated by the name, is conceptually based on the natural process of solids cooling down.
When a solid cools, it reaches an equilibrium at each temperature.
Simulated annealing takes advantage of this natural procedure to find the global optimum of a cost function that has many local optima, and does not depend on the quality of the initial solution~\cite{van1987simulated, szu1987fast, aarts1988simulated, romeijn1994simulated}.
In this method, the algorithm generates candidate solutions that are nearby the current solution.
The method accepts the candidate solution if it is better than the current solution.
If the candidate solution is worse, however, the algorithm may still accept the candidate solution with some probability.
This enables the algorithm to explore sub-optimal regions of the parameter space, helping the algorithm to escape local optimums \cite{romeijn1994simulated}.
This process is repeated while time allows or until convergence, producing near-optimal solutions \cite{aarts1988simulated, romeijn1994simulated}.

Simulated annealing has been used for a variety of different layout planning problems \cite{serafini1994simulated ,mckendall2006simulated, csahin2011simulated, ahonen2014simulated, palubeckis2015fast}.
In 2001, Baykasoglu et al. \cite{baykasouglu2001simulated} demonstrated the applicability of simulated annealing in dynamic manufacturing facility layout planning.
The algorithm was further developed and used in more complex static and dynamic layout planning problems with multiple objective functions such as corridor allocation\cite{ahonen2014simulated}, arranging manufacturing facilities\cite{mckendall2006simulated}, and single-row equidistant facility layout problems~\cite{palubeckis2015fast}.

Despite much research in automatic industrial and commercial layout planning, optimizing the layout of the interior of hospital rooms considering patient safety has not yet been studied and is the subject of this work.

\vspace{-5pt}
\section{\uppercase{Method}}\label{section: method}
\noindent Our method optimizes the interior layout of a hospital room to create a safer environment for the patient with respect to risk of fall.
To do so, we define a cost function
built around the fall risk assessment model developed in \cite{sabbagh2020development}.
We minimize this cost function using simulated annealing and leverage real-world architectural design guidelines~\cite{ramsey2007architectural, neufert2012architects} as constraints to ensure room functionality.

\subsection{Hospital Room Layout}

\noindent A typical single-bed hospital room consists of two sub-rooms, a main room, where the patient, visitor and clinical zones are located, and a bathroom where the toilet, shower and sink are located. The geometry of the room boundaries are determined via architectural considerations and are inputs to our method (See Fig.~\ref{fig:first} for the room geometry used in this paper).

For each of these sub-rooms, specific furniture items must be placed, light sources included to illuminate the room, and doorways must connect the bathroom to the main room and the main room to the hallways.
In this work, we optimize the placement of the furniture, light sources, and doorways to lower the risk of patient fall.

Borrowing notation from the constraint satisfaction problem literature, we formalize the input to our method as three sets:
$\X = \{x_0, x_1, \dots, x_n\}$, a set of $n$ variable objects (e.g., furniture, lights sources, and doors) for the room; $\D = \{D_0, D_1, \dots, D_n\}$, a set of domains that are defined for each variable in $\X$; and $\C$, a set of constraints defined over the variables in $\X$, where each constraint may relate any subset of the variables.

For each object $x_i$ in $\X$, we parameterize its placement in the hospital room as the configuration vector $\vec{d}_i \in D_i$.
We then parameterize the layout of the entire hospital room as the vector $\layout = [\vec{d}_0^T, \vec{d}_1^T, \dots, \vec{d}_n^T]^T$, the concatenation of the configuration vectors for each of the objects in $\X$.

At a high level, the goal of the method is to determine a specific layout $\layout$ of the hospital room that respects each constraint in $\C$, while minimizing a function that relates $\layout$ to the risk of patient falls.

\subsection{Quantifying Fall Risk} \label{subsection: cost function and fall risk}

\noindent To evaluate a specific room layout we build on the fall risk assessment model proposed in~\cite{sabbagh2020development} in which the overall fall risk distribution of a room is calculated as a function of a set of \
factors extrinsic to the patient taken from previous studies of hospital fall risk. 
This fall risk model considers both static and dynamic factors affected by the room configuration and provides two levels of fall risk evaluation: (1) A room baseline evaluation that is calculated solely from the static factors of: floor type, lighting condition, door operation, and the supporting or hazardous effect of the surrounding objects (e.g. grab bars, chair, medical stands, sofa, sink, toilet, and bed) resulting in a risk distribution over the entire room; and (2) Motion-based evaluation that considers patient ambulation defined by dynamic gait properties such as the turning angle and the type of activity such as sit-to-stand, walk, turn, and stand-to-sit.

The input to this risk model includes details about the room such as floor surface type as well as the room layout defining the lighting, locations and configurations of all objects, and door placements.
The output is a risk distribution defined as a value for each element of a grid map discretizing the hospital room, $r(\layout)$.

In the baseline layer of the model, each grid has a base value of 1. Then, the value is modified based on the distance to the closest supporting object as well as lighting, flooring, and door operation factors.
For the motion-based evaluation, the model uses a set of pre-defined scenarios such as the patient's transitions from bed to toilet and then predicts sample trajectories between objects for each scenario. 
Each grid cell through which the simulated trajectory passes is influenced by the specific simulated activity, such as sit-to-stand, and dynamic factors such as angular velocity and turning angle.

As multiple trajectories are possible between two target objects, the model generates and evaluates a distribution of simulated patient trajectories.
The distribution of the fall risk for the entire room is obtained by combining the baseline and motion-based evaluation risk profiles.
To do so, the average of the baseline fall risk factors and the fall risk of the points of the trajectories/activities laying on each grid cell is calculated. The final output is a risk value for each grid cell.
This can be visualized as a heat map over the hospital room layout showing the distribution of fall risk values (e.g., see Fig.~\ref{fig:first}).
These values correspond to the percentage of increase or decrease in the risk of fall for each grid cell. Values greater than 1.0 (red cells) mean that there is high fall risk, and values less than 1.0 (blue cells) show that the fall risk is decreased. See \cite{sabbagh2020development} for more details on the specific framework and contributing factors.

In our work we optimize a cost function designed for this set of fall risk values that weighs different aspects of the distribution, as there are multiple potential aspects of the distribution that are relevant.
For example, if we define the cost function as the maximum fall risk in the room, the optimizer may focus on reducing the risk of fall of the single worst grid cell while the fall risk of the rest of the room remains unacceptably high.
On the other hand, if we define the fall risk as the mean or median fall risk over all the grid cells of the room, the algorithm may reduce the mean or median while a few grid cell values may remain extremely high.
Further, it may be the case that the width of the high-risk tail of the distribution is an important factor in assessing the overall risk of a room, as it represents a set of high-risk areas.
Each of these considerations relates the fall risk distribution to a different objective. Thus, we define a weighted combination of the median and the maximum of the fall risk distribution, as well as a metric of the area under the high-risk tail of the distribution as an aggregate representation of fall risk.
The specific weighting of these considerations can then be set by the user based on their specific, clinically-motivated preferences.
In this work we treat this weighting as input to our method.

\begin{figure}
\includegraphics[width=\columnwidth]{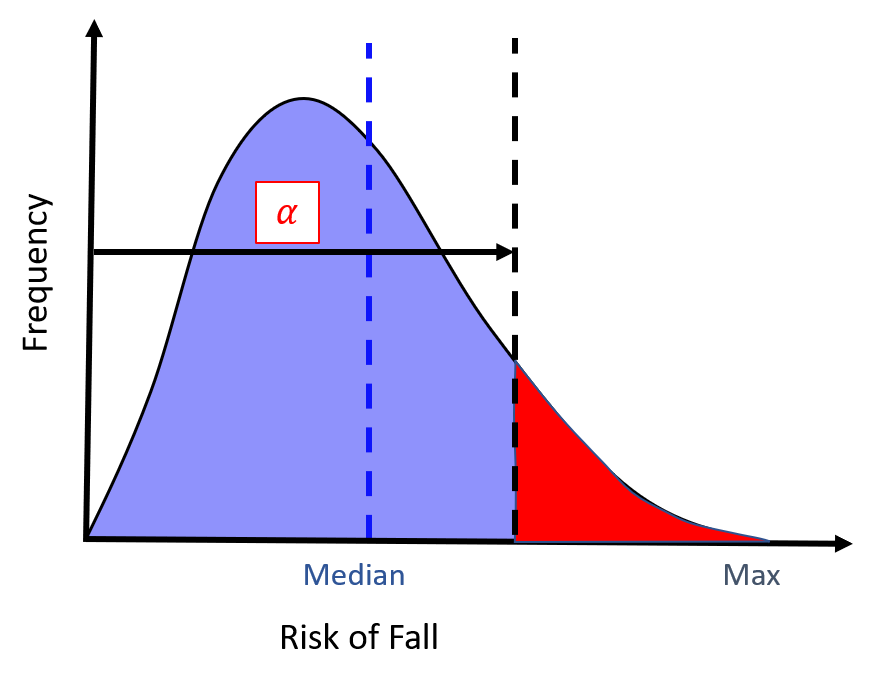}
\caption{\footnotesize Cost function parameters of the fall risk distribution, showing the median, maximum, and the area under the high-risk tail. Parameter $\alpha$ is the cut-off value for the area under the curve and the red region shows the area considered by the cost function.}
\label{fig: costFunctionParameters}
\end{figure}

Specifically, we define the cost function $f(\layout)$ as:
\begin{equation}
\begin{split}
f(\layout) =& \omega_1 \textrm{median}(r(\layout)) + \omega_2 \textrm{max}(r(\layout)) \\
&+ \omega_3 \left(\frac{\alpha-\textrm{mean}(r(\layout))}{\textrm{std}(r(\layout))}\right)    
\end{split}
\label{eq: final cost func}
\end{equation}
where $\alpha$ is a user specified cut-off parameter that determines where in the distribution to begin considering the area under the tail, as seen in Fig.~\ref{fig: costFunctionParameters}.
The third term in Eq.~\ref{eq: final cost func} represents the aggregation of the grid cells with risk values higher than $\alpha$.
The concept of considering the area under the curve tail is a familiar concept in finance risk management and is known as the conditional value-at-risk~\cite{rockafellar2002conditional}.
Our final cost function is then a function of the median, mean, standard deviation, and maximum value of the fall risk distribution.
Due to the discretization of the room and the random simulation of patient movement in $r(\layout)$, along with other factors, $f(\layout)$ is both highly non-linear and not differentiable.
This motivates our use of a gradient-free optimization method when optimizing $f(\layout)$.

\subsection{Optimizing the Layout} \label{subsection: sa}
\noindent In this work we leverage simulated annealing for optimizing $f(\layout)$ as it is a well known stochastic iterative optimization method for gradient and derivative free cost functions such as $f(\layout)$.
At each iteration, a layout nearby the current layout is generated at random and its cost is evaluated.
If it is an improvement over the current layout it is accepted and becomes the current layout for the next iteration.
If it is worse, it may still be accepted based on the Metropolis Probability~\cite{aarts1988simulated, van1987simulated, metropolis1953equation}:
\begin{equation} \label{eq: boltzman probability}
P_\textrm{Metropolis} = \exp(\frac{-\Delta \textrm{c}}{\kappa T})  
\end{equation}
where $\kappa$ is the Boltzman constant and $T$ is a ``temperature" value which decays over time according to a cooling schedule and $c$ stands for the cost value associated with the layouts.

The temperature scheduling function we use in our method is widely used and was first introduced by Kirkpatrick et al. in 1983 \cite{kirkpatrick1983optimization},
\begin{equation}\label{eq: temp scheduling}
T_i = k T_{i-1}
\end{equation}
where $T_i$ is the temperature at cycle $i$ and is based upon the previous cycle's temperature, $T_{i-1}$, and $k$ is a constant factor $(0<k<1)$ controlling the rate of temperature decay~\cite{faber2005dynamic}.

The algorithm, detailed in Alg.~\ref{alg: SA}, begins with an initial layout.
It then improves upon this layout for a set number of cycles, where the temperature is decreased according to Eq.~\ref{eq: temp scheduling} between each cycle.
Within each cycle, the algorithm generates candidate layouts, $\layout_\textrm{next}$ during a number of trials.
In each trial, the algorithm checks if the new candidate layout's cost value, $c_\textrm{next}$ is lower than the previously accepted layout's cost $c_\textrm{current}$.
If so, the system accepts the candidate layout for the next trial.
However, if the cost value of the candidate layout is higher than the previously accepted solution, the algorithm may still accept it depending on the acceptance probability defined by Eq. \ref{eq: boltzman probability}, enabling the algorithm to avoid local minima.
As the algorithm progresses this acceptance probability decreases.

\begin{algorithm}
\caption{Simulated Annealing for Hospital Room Optimization}\label{alg: SA}
\begin{algorithmic}[1]
\State \textbf{Initialization:}
\State \hspace{\algorithmicindent}$T_0 \leftarrow$ initial temperature
\State \hspace{\algorithmicindent}$k \leftarrow \text{temperature decreasing factor}$
\State \hspace{\algorithmicindent}$\kappa \leftarrow \text{Boltzman constant}$
\State \hspace{\algorithmicindent}$\vec{\sigma}_r \leftarrow \text{vector of standard deviations}$
\State \hspace{\algorithmicindent}$\text{numCycle}\leftarrow \text{number of cycles}$
\State \hspace{\algorithmicindent}$ \text{numTrial}\leftarrow \text{number of trials per cycle}$
\State \hspace{\algorithmicindent}$ \layout_0\leftarrow \text{initial room layout}$
\State \hspace{\algorithmicindent}$ \layout_\mathrm{current}\leftarrow\layout_0$ \Comment{the current layout}
\State \hspace{\algorithmicindent}$ \layout_\mathrm{best}\leftarrow \layout_0$ \Comment{the best layout found}
\State \hspace{\algorithmicindent}$c_\mathrm{current}\leftarrow f(\layout_0)$ \Comment{cost of current layout}
\State \hspace{\algorithmicindent}$c_\mathrm{best}\leftarrow f(\layout_0)$ \Comment{cost of best layout found}
\For {$i = 1$ to numCycle}
\State $T_i = k T_{i-1}$
    \For {$j = 1$ to numTrial}
    \State $\layout_\mathrm{next} \leftarrow \texttt{NearbyFeasLayout}(\layout_\mathrm{current}, \vec{\sigma}_r)$
    \State $c_\mathrm{next}\leftarrow f(\layout_\mathrm{next})$
    \State $P = \exp(\frac{c_\mathrm{next}-c_\mathrm{current}}{\kappa T_i})$
    \State $r \leftarrow$ generate a random number in $(0,1)$
    \If {$c_\mathrm{next} < c_\mathrm{current}$ \textbf{or} $r < P$} 
        \State $\layout_\mathrm{current} \leftarrow \layout_\mathrm{next}$
        \State $c_\mathrm{current} \leftarrow c_\mathrm{next}$
        \If{$c_\mathrm{current} < c_\mathrm{best}$}
            \State $\layout_\mathrm{best} \leftarrow \layout_\mathrm{current}$
            \State $c_\mathrm{best} \leftarrow c_\mathrm{current}$
        \EndIf
    \EndIf
        
    \EndFor
\EndFor \\ 
\Return $\layout_\mathrm{best}, c_\mathrm{best}$
\end{algorithmic}
\end{algorithm}

In this way, the method generates better and better layouts over time, iteratively improving the layouts with respect to our cost function, returning the best layout found at the conclusion of the method's execution.

\subsection{Generating Feasible Layouts} \label{subsection: room generation}

\noindent When generating both a random initial feasible room layout (Alg.~\ref{alg: SA} line 8) and nearby feasible layouts (Alg.~\ref{alg: SA} line 16, $\texttt{NearbyFeasLayout}$) during the execution of the optimization, it is important that the layouts are \emph{feasible}, i.e., satisfy all of the constraints in the constraint set $\C$.
In both cases, we employ a random sampling with backtracking approach.
During the generation of the initial room layout we sample uniformly at random from each variable's domain. When generating nearby layouts to an existing layout, we sample each variable's configuration from a normal distribution centered around the existing layout's configuration and with standard deviation defined for each variable ($\vec{\sigma}_r$ in Alg.~\ref{alg: SA}).
\begin{figure}
	\vspace{5pt}
    \centering
    \begin{subfigure}[t]{0.28\textwidth}
        \includegraphics[width=\textwidth]{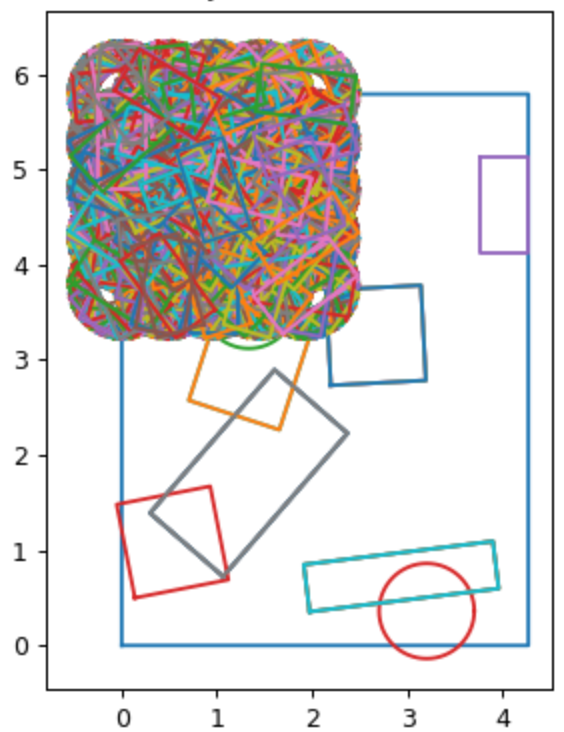}
        \caption{Unsuccessful object placement.}
        \label{subfig: Unsuccessful Object Placement}
        \vspace{5pt}
    \end{subfigure}\qquad
    \begin{subfigure}[t]{0.28\textwidth}
        \includegraphics[width=\textwidth]{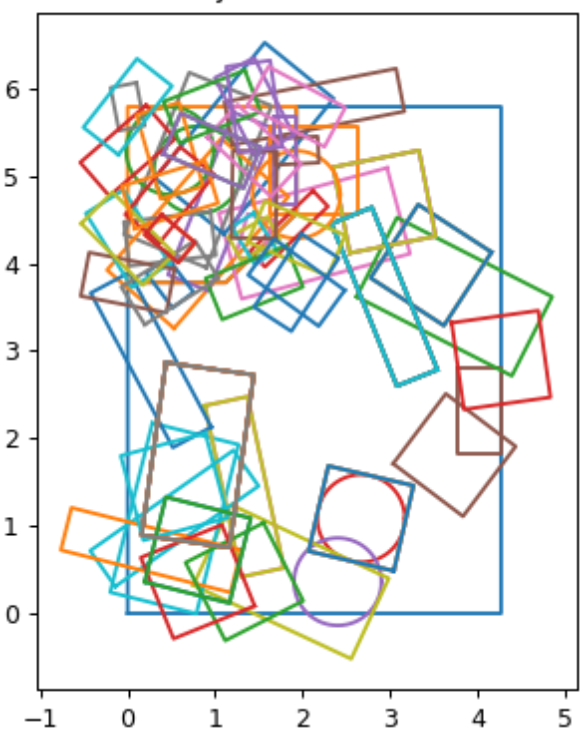}
        \caption{Multiple object placement trials}
        \label{subfig: Multiple Object Placement Trials}
        \vspace{5pt}
    \end{subfigure}
    \begin{subfigure}[t]{0.28\textwidth}
        \includegraphics[width=\textwidth]{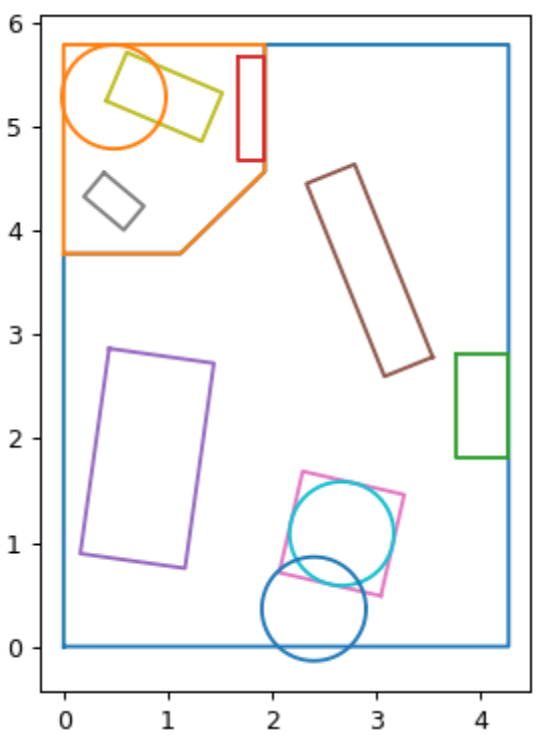}
        \caption{Final object placement}
        \label{subfig: final object placement}
        \vspace{5pt}
    \end{subfigure}
    \caption{Object placement procedure. (a) Previously placed objects prevent feasible placement of a subsequent object. (b) Backtracking changes the placement of previously placed objects. (c) All objects are placed in a way that satisfies the constraints.}\label{fig: object placement}
\end{figure}
To ensure the constraints are satisfied during this process we leverage a backtracking search method \cite{karumanchi2018algorithm}.
Backtracking chooses values for one variable at a time, checking for constraint satisfaction as variables are assigned.
The method re-samples when a variable assignment violates constraints, and backtracks to re-assign previous variables when a variable has no legal values left to assign, or when a maximum number of iterations are reached or a maximum time has elapsed while attempting to assign that variable.
In this way, backtracking recursively returns to previously assigned values, changing their assignments to satisfy the constraints.

\begin{figure*}
\includegraphics[width=\textwidth]{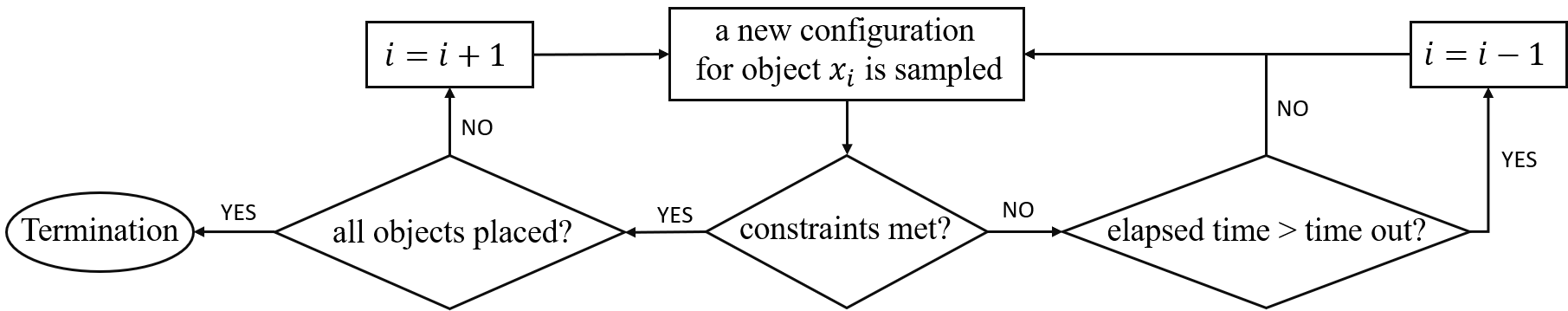}
\caption{\footnotesize A flowchart of the backtracking method}
\label{fig: backtracking}
\end{figure*}

Figure \ref{fig: object placement} shows an example of backtracking during the placement of an object in the hospital room.
In Fig. \ref{subfig: Unsuccessful Object Placement}, the algorithm attempts to place one of the objects that is required to be in the bathroom, but because of the placement of the previous objects, it was not able to do so.
The method backtracks and considers other placements for previously placed objects (Fig \ref{subfig: Multiple Object Placement Trials}).
Finally, it successfully places the objects in the room while satisfying the constraints (Fig \ref{subfig: final object placement}).
Figure \ref{fig: backtracking} depicts the overall flow of the backtracking algorithm.

The constraints in $\C$ ensure that the hospital room layouts generated by our method maintain functionality based on architectural regulation.
The constraints depend on the objects to be placed within the room and can be defined over a single object, such as requiring certain types of furniture to be adjacent to a wall and requiring a light to be placed in the boundaries of the room, or multiple objects such as ensuring that there exists a minimum clearance on both sides of a bed.
Our method takes the constraint set as an input and makes no assumptions about the properties of the constraint functions, such as differentiability, other than to require that the constraint functions return a boolean value indicating whether a specific layout violates or satisfies the constraints.
The specific constraints we use in our experiments are described in Sec.~\ref{sec: implementation and experimental result}.

\section{\uppercase{Implementation and Experimental Results}} \label{sec: implementation and experimental result}
\noindent To assess the performance of the proposed model, we optimize the interior configuration of two common typologies of hospital rooms: inboard rooms and outboard rooms.
In the inboard architecture, the bathroom is located near the entry of the room and next to hallways.
In the outboard room, the bathroom is placed along the exterior wall of the room (see Fig.~\ref{fig: inOutBoardWalls}).
In this paper, we demonstrate the efficacy of our optimization method on a representative sample room from each of these room types using dimensions and shapes used in real hospitals.

\subsection{Implementation Details}

\begin{figure}[t]
\includegraphics[width=\linewidth]{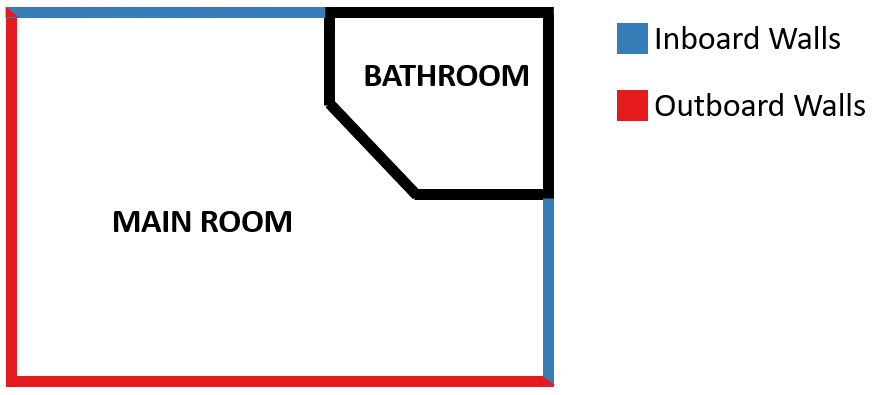}
\caption{\footnotesize Schematic of the hospital room. Blue lines are permissible hallway door placements for outboard rooms and red lines show the permissible hallway door placement walls for inboard rooms.}
\label{fig: inOutBoardWalls}
\end{figure}

\noindent For both the inboard and outboard room experiments we define $\X$ to include: furniture consisting of a sofa, the patient bed, a patient chair, a visitor chair, a mobile medical stand, a toilet, and a sink; a ceiling light for both the main room and the bathroom; and a door connecting the bathroom to the main room and the main room to the hallway (see Fig.~\ref{fig: inOutBoardWalls}).

\begin{figure*}[h]
\includegraphics[width=\textwidth]{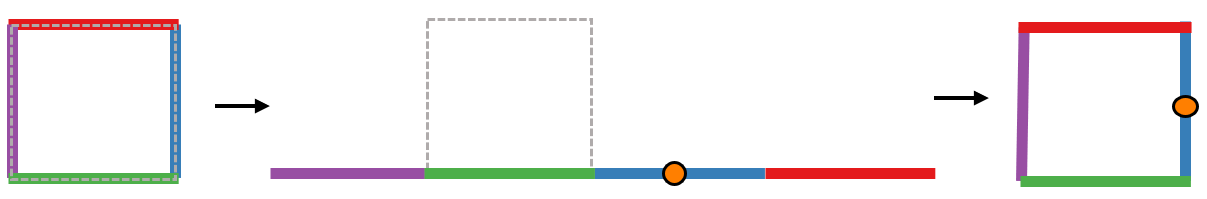}
\caption{\footnotesize Sampling a point on the walls of the room for objects that are constrained to be against the walls. We unwrap the walls of the room into a single line, sample a point on the line and then find the coordinates of the sampled point in the original space with the walls in their original 2D geometry.}
\vspace{20pt}
\label{fig: unwrapWall}
\end{figure*}

We make a distinction between the furniture objects that are required to be placed against the wall for functionality (the patient bed, sofa, sink, and toilet) and the furniture objects that are allowed to be placed freely throughout the inside of the room (the patient chair, visitor chair, and mobile medical stand).
Each of the furniture objects that can be placed freely inside the room have domains that include two position values, representing the x and y location, and an orientation value, i.e., $\Reals^2 \times (S)^1$.
The lights have domains that include the position values but not orientation, $\Reals^2$.
For both the furniture objects required to be placed against the wall and the doors we implicitly represent this using a domain defined by a single real value associated with the object's location along a parameterized representation of the walls unwrapped as a line (See Fig.~\ref{fig: unwrapWall} for an example).

We define our constraints based on architectural guidelines\cite{ramsey2007architectural, neufert2012architects}.
We require certain objects to be placed in specific sub-rooms as well as clearance values around certain types of objects.
These are detailed in Table~\ref{tab: constraints}.
The clearance value constraints are designed to ensure functional use by an average size adult, however we recognize that additional considerations such as those defined by the American's with Disabilities Act (ADA) may have different requirements.
Further, we require a feasible layout to have one light in the bathroom area and one light in the main room.
We require the bathroom door to connect the bathroom and main room and the main door to connect the main room and the hallway.
We also require that no objects' geometries overlap in the $x,y$ plane, with the exception of the lights which are placed on the ceiling and as such do not collide with objects placed on the floor.
Moreover, the algorithm places each object in the correct sub-room (see Table~\ref{tab: constraints}).
For example, the bed must be placed in the main room.
For objects that can be positioned in either room, such as the cabinets or sink, the user specifies in which room the object should be placed.

\begin{table*}[hbt!]
  \centering
  \begin{tabular}{||c||c|c|c|c|c||}
  \hline
  Object & Bed & Sofa \& Chairs & Toilet & Sink\\
     \hline
    Clearance constraint & 0.4m, both sides  & 0.35m, front & 0.4m, front& 0.35m, front \\
    Sub-room constraint & Main room  & Main room & Bathroom & Bathroom \\
    \hline
\end{tabular}
  \caption{Clearance constraints for objects placed in the rooms along with the rooms the object belongs to.}
  \label{tab: constraints}
\end{table*}

The difference between the outboard and inboard room typology manifests in the constraint associated with the door connecting the patient living area to the hallway.
In an inboard room, this door must be placed on one of the two walls shared by the bathroom, and in an outboard room it must be placed on one of the two walls not shared by the bathroom (see Fig.~\ref{fig: inOutBoardWalls}).

The trajectories that we use when calculating the fall risk (see Sec.~\ref{subsection: cost function and fall risk}) are: ambulation from the bed to the patient chair, from the bed to the toilet, and from the bed to the main door.
We set the parameters in Eq.~\ref{eq: final cost func} with $\alpha$ chosen to be $95\%$ of the $\textrm{max}(r(\layout))$ value and $\omega_1 = \omega_2 = \omega_3 = 0.33$.
\begin{table*}[hbt!]
  \centering
  \begin{tabular}{||c||c|c|c|c|c|c|c|c||}
  \hline
  Objects &  \multicolumn{4}{|c|}{Main room} & \multicolumn{4}{|c|}{Bathroom}\\
  \cline{2-9}
     & $\sigma_x$&$\sigma_y$ &$\sigma_{\theta} $& $\sigma_w $&$ \sigma_x $&$ \sigma_y $&$ \sigma_{\theta} $&$ \sigma_w $\\
     \hline
    Furniture (generic) & 1m & 1m& $30^{\circ}$ & - &0.5m & 0.5m & $30^{\circ}$ & - \\
    Furniture (against wall) & - & - & - & 5m & - & - & - & 1m \\
    Lights & 1m & 1m & - & - & 1m & 1m & - & - \\
    Doors & - & - & - & 4m & - & - & - & 2m \\
    \hline
\end{tabular}
  \caption{Values of $\vec{\sigma}_r$. Parameters $\sigma_x$ and $\sigma_y$ are the position of the object inside the room. $\sigma_{theta}$ stands for the orientation values. Parameter $\sigma_w$ shows the standard deviation value for the objects that are attached to a wall.}
  \label{tab: initial values for sigma}
\end{table*}
The initial temperature value in our simulated annealing is chosen to be 10, $k$ is set to $0.8$, $\kappa$ is set to 1, numCycles is set to 25, and numTrials is set to 30 based on a heuristic analysis of our cycles' acceptance rates (as in~\cite{szu1987fast} and~\cite{duque1997constructing}).
The values in $\vec{\sigma}_r$ depends on the specific object and their corresponding sub-room.
Table \ref{tab: initial values for sigma} shows the set values for $\vec{\sigma}_r$.
During initial and near-by room generation, our timeout to initiate backtracking is set at $5$ seconds.

For both the inboard and outboard room typologies we use the room geometry shown in Figs~\ref{fig:first}, ~\ref{fig: outboard room optimization with HM}, and~\ref{fig: inboard room optimization with HM}.

\section{Results and Discussion}
\begin{figure}
\includegraphics[width=\linewidth]{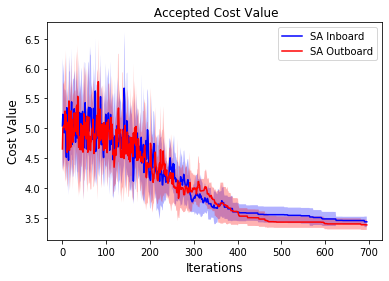}
\caption{\footnotesize The cost value for the layout during the optimization iterations for the inboard room type (blue) and the outboard room type (red). The shaded regions show the corresponding standard deviation of each cost value.
}
\label{fig: comparison accepted}
\end{figure}

\noindent We studied the performance of our optimization algorithm in reducing the risk of fall in two room typologies. 
We perform ten optimization runs for each room typology.
The average time required for each optimization run was $(2.4359\pm 0.8453) \times 10^4$ seconds ($\approx 7$ hours).
In Fig.~\ref{fig: comparison accepted} we plot the cost value of the layout being considered at each iteration (each trial of each cycle), i.e., $c_\mathrm{current}$, averaged over the ten runs.
As can be seen, the simulated annealing algorithm rapidly explores the parameter space early in the runs, escaping many local minima, however as the run progresses the algorithm settles upon low cost layouts.

\begin{figure}
\includegraphics[width=\linewidth]{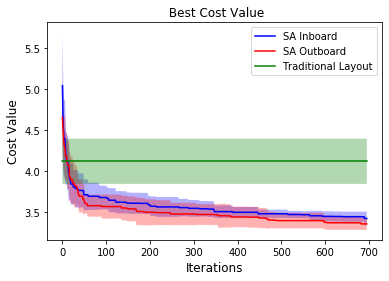}
\caption{\footnotesize The cost value for the best layout found up to a given iteration under our method for the inboard room type (blue) and the outboard room type (red). We also plot the value associated with the traditional layout for reference. The shaded regions represent the corresponding standard deviation.}
\label{fig: comparison best}
\end{figure}

\begin{figure}
    \centering
    \begin{subfigure}{0.2\textwidth}
        \includegraphics[width=\textwidth]{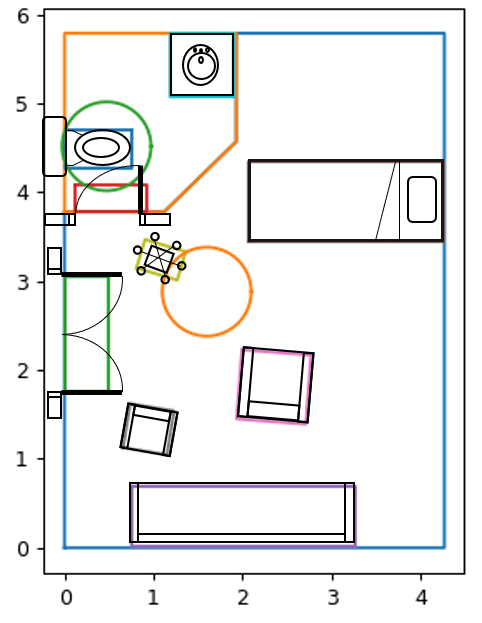}
        \caption{Initial room layout}
        \label{subfig: init inboard room}
    \end{subfigure}
    \begin{subfigure}{0.2\textwidth}
        \includegraphics[width=\textwidth]{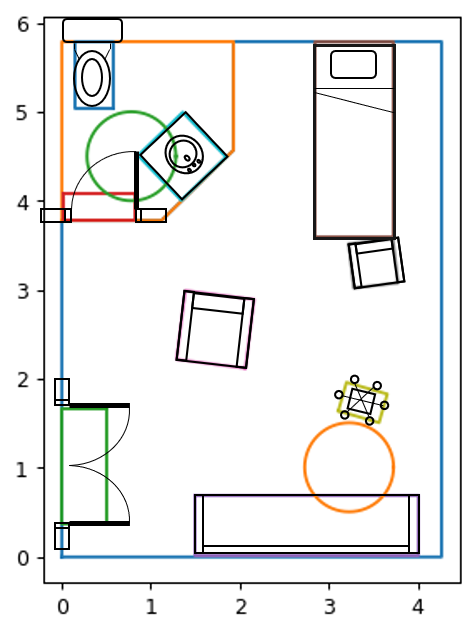}
        \caption{Final room layout}
        \label{subfig: final inboard room}
        \vspace{10pt}
    \end{subfigure}\qquad

    \begin{subfigure}{0.2\textwidth}
        \includegraphics[width=\textwidth]{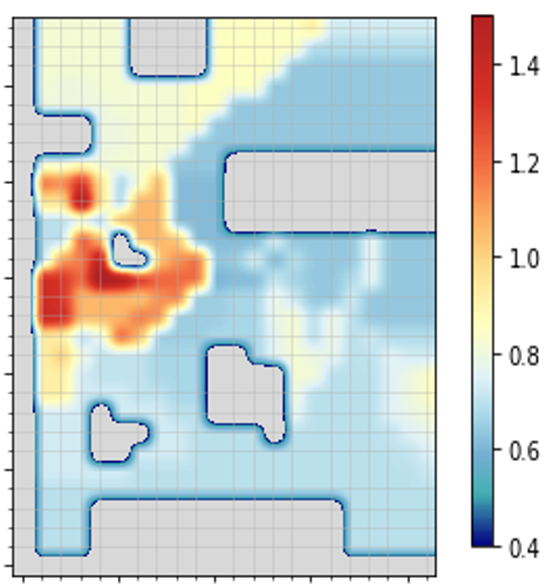}
        \caption{ Initial risk heatmap}
        \label{subfig: init inboard HM}
    \end{subfigure}
    \begin{subfigure}{0.2\textwidth}
        \includegraphics[width=\textwidth]{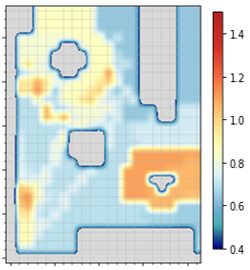}
        \caption{ Final risk heatmap}
        \label{subfig: init inboard HM}
    \end{subfigure}
    \vspace{5pt}
    \caption{Inboard room layout evaluation and optimization with respect to fall risk. (a) and (b) show schematics of the inboard rooms generated by the algorithm. (a) is an instance of the initial room layout and (b) shows the optimized room layout for one of the runs. (c) and (d) show the corresponding heat map of the risk of fall as evaluated by the fall risk model. Higher values on the color-bars of figures (c) and (d) indicate higher risk of falls.}\label{fig: inboard room optimization with HM}
\end{figure}

In Fig.~\ref{fig: comparison best} we plot the cost value of the best layout found up until that iteration in the optimization, i.e., $c_\mathrm{best}$, averaged over the ten runs.
The average starting cost value for the randomly initialized inboard rooms and outboard rooms were $5.04\pm 0.4$ and $4.67\pm 0.65$, respectively.
After optimization, the lowest cost value for the generated rooms for the inboard type was $3.42\pm 0.08$ and $3.36\pm 0.07$ for the outboard type.
This represents a reduction of $40\%$ for the inboard room and $42\%$ for the outboard room.
We also evaluated the risk-of-fall-based cost value of a traditional hospital room layout, as depicted in Fig.~\ref{subfig: trad room}.
The cost value of the traditional layout was found to be $4.1\pm 0.27$.
Compared with the traditional layout, our optimized layouts achieve a cost reduction of $18.05\%$.
To evaluate the statistical difference between the traditional room cost value, and the starting cost values and the final cost values for the generated room typologies, we applied the two sample Kolmogorov-Smirnov test (K-S test), a statistical test designed to determine if two continuous or discrete distributions are significantly different \cite{chakravarti1967handbook}.
We examined the similarity between the distributions in five cases: 1) inboard: the cost values of the initial randomized layouts vs the cost values of the final layouts, 2) outboard: cost values of the initial randomized layouts vs cost values of the final layouts, 3) the final cost values of the optimized inboard rooms vs the final cost values of the optimized outboard rooms, 4) the final cost values of the optimized inboard rooms vs the cost values of the traditional room, and 5) the final cost values of the optimized outboard rooms vs the cost values of the traditional room.
In all five cases, the null hypothesis was rejected at the $5\%$ significance level.
For the initial vs final cost values for both the inboard and outboard rooms, and the comparison between the optimized inboard and outboard rooms with the traditional room, the order of the asymptotic $p$-value was $10^{-5}$ with K-S score of 1.
In rejecting the null hypothesis for inboard vs outboard final cost values, the $p-$value was computed as 0.0310 with a K-S score of 0.6, which indicates more similarity than between the initial and final layouts, but still significantly different.
\begin{figure}[]
    \centering
    \begin{subfigure}[t]{0.2\textwidth}
        \includegraphics[width=\textwidth]{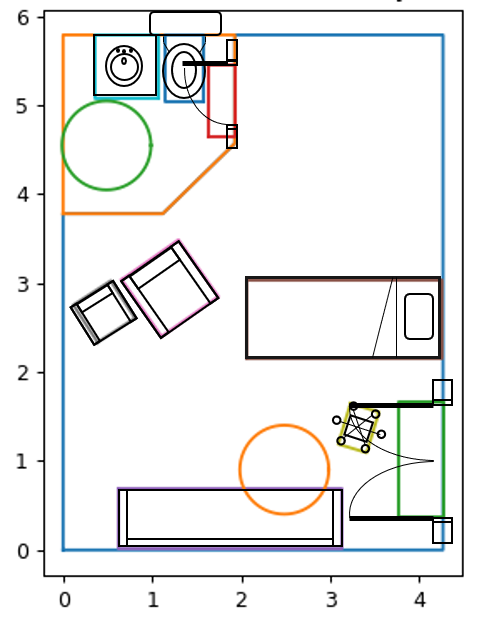}
        \caption{Initial room layout}
        \label{subfig: init inboard room}
    \end{subfigure}
    \begin{subfigure}[t]{0.2\textwidth}
        \includegraphics[width=\textwidth]{Figures/outboard_room_best_objects.png}
        \caption{Final room layout}
        \label{subfig: outboard- final inboard room}
        \vspace{10pt}
    \end{subfigure}\qquad

    \begin{subfigure}[t]{0.2\textwidth}
        \includegraphics[width=\textwidth]{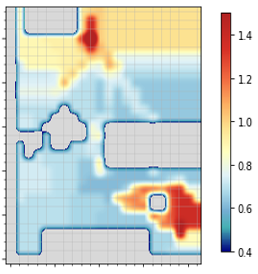}
        \caption{Initial risk heatmap}
        \label{subfig: outboard-init inboard HM}
    \end{subfigure}
    \begin{subfigure}[t]{0.2\textwidth}
        \includegraphics[width=\textwidth]{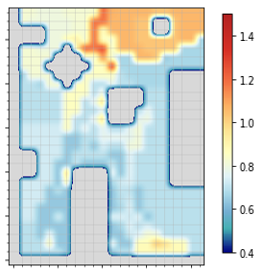}
        \caption{ Final risk heatmap}
        \label{subfig: outboard-init inboard HM}
    \end{subfigure}
    \vspace{5pt}
    \caption{Outboard room layout evaluation and optimization with respect to fall risk. (a) and (b) show schematics of the outboard rooms generated by the algorithm. (a) is an instance of the initial room layout and (b) shows the optimized room layout for one of the runs. (c) and (d) show the corresponding heat map of the risk of fall as evaluated by the fall risk model. Higher values on the color-bars of figures (c) and (d) indicate higher risk of falls. }\label{fig: outboard room optimization with HM}
\end{figure}
In Figs.~\ref{fig: outboard room optimization with HM} and ~\ref{fig: inboard room optimization with HM}, we show representative inboard and outboard room layouts, both before and after optimization, as well as their associated fall risk score heatmaps.
In Fig.~\ref{fig:first} we show the same for the traditional layout.
One potentially interesting trend is the lower cost values found when optimizing the outboard rooms compared with the inboard rooms.
Looking at Fig. \ref{fig: inOutBoardWalls}, placing the hallway door on one of the further walls to the bathroom will place the door in between the bed and the bathroom. This results in trajectories with fewer sharp turns, which decreases the risk of falls as defined in the fall risk model.
This may also result from outboard walls being longer than the inboard walls for our room geometry, and hence, the algorithm has more options of sampling points for door placement and potentially more configuration options.

\section{\uppercase{Conclusion and future work}}
\noindent In this work, we presented a method that built upon a patient fall-risk model and presented a gradient-free optimization method, based on simulated annealing, to reduce the risk of patient falls in hospital rooms by optimizing the configurations of objects inside the room.
We evaluated our method on two room typologies, inboard and outboard, with common hospital room objects utilized in both.
The algorithm optimized the layout of the rooms with respect to a cost function that was defined based on the distribution of the patient fall risk in the rooms, which considered both static factors associated with the object placements in the room as well as the kinematics and dynamics of simulated patient trajectories.

Our method leverages constraints based on object functionality and architectural guidelines used for facility layout planning.
We demonstrated results, averaging ten runs for each room type, showing significant improvement with respect to our patient fall-risk cost metric compared to both random initial room layouts and traditional hospital room layouts.

We chose simulated annealing as a canonical optimization option for complex gradient-free problems such as ours and intend the results to demonstrate the feasibility of utilizing optimization in the problem domain.
However we recognize that many other optimization methods may be applicable and intend to investigate the use of other methods such as genetic algorithms, particle swarms, gray wolf optimizer, etc, in the future.
Further, we intend to investigate the use of other metrics defined over the fall-risk distribution beyond the weighted metric presented in this work.
We will incorporate expert feedback from hospital designers, architects, and healthcare providers into the optimization loop.
This feedback will be used to enhance the fall risk model and optimization method to improve our result and achieve a functional, safe room layout.

We will also conduct human subject studies to evaluate the result of optimized room layouts in reducing the risk of patient falls in a simulated hospital room.
While our method and the fall-risk model it builds upon are specific to hospital room layouts, our method may have applications in reducing falls in at-risk populations outside of healthcare settings as well.
These other environments include assisted living, and long-term care facilities, and homes of individuals at high risk of falls.

We believe that this work takes significant steps toward demonstrating the feasibility of optimizing the layout of hospital rooms in order to reduce the risk of patient falls and improve patient outcomes.

\section*{\uppercase{Acknowledgment}}
This project was supported by grant number R18HS025606 from the Agency for Healthcare Research and Quality (AHRQ). The content is solely the responsibility of the authors and does not necessarily represent the official views of the Agency for Healthcare Research and Quality.

\bibliographystyle{apalike}
{\small
\bibliography{references}}

\end{document}